\documentclass[twocolumn, switch]{article} % Method A for two-column formatting

\usepackage{preprint}
\usepackage{algorithm}
\usepackage{algorithmic}
\usepackage{amsmath, amsthm, amssymb, amsfonts}

\usepackage[numbers,square]{natbib}
\usepackage[utf8]{inputenc}	% allow utf-8 input
\usepackage[T1]{fontenc}	% use 8-bit T1 fonts
\usepackage{xcolor}		% colors for hyperlinks
\usepackage[colorlinks = true,
            linkcolor = purple,
            urlcolor  = blue,
            citecolor = cyan,
            anchorcolor = black]{hyperref}	% Color links to references, figures, etc.
\usepackage{booktabs} 		% professional-quality tables
\usepackage{threeparttable}	% tables with footnotes
\usepackage{nicefrac}		% compact symbols for 1/2, etc.
\usepackage{microtype}		% microtypography
\usepackage{lineno}		% Line numbers
\usepackage{float}			% Allows for figures within multicol

\usepackage{lipsum}		%  Filler text

\usepackage{newfloat}
\DeclareFloatingEnvironment[name={Supplementary Figure}]{suppfigure}
\usepackage{sidecap}
\sidecaptionvpos{figure}{c}

\usepackage{titlesec}
\titlespacing\section{0pt}{12pt plus 3pt minus 3pt}{1pt plus 1pt minus 1pt}
\titlespacing\subsection{0pt}{10pt plus 3pt minus 3pt}{1pt plus 1pt minus 1pt}
\titlespacing\subsubsection{0pt}{8pt plus 3pt minus 3pt}{1pt plus 1pt minus 1pt}

\title{Systematic Lightweight Method for Robotics Based on Strain Energy Distribution Optimization}

\usepackage{titling}
\usepackage{orcidlink}
\usepackage{footmisc}
\newcommand{\Author}[3]{% Name, ORCID, Institution
  \textbf{#1}\textsuperscript{#2},\ \orcidlink{#3} %
}

\author{
  \Author{Jingchen Li}{1}{0000-0003-3196-7882}
}

\date{%
  \textsuperscript{1}Tencent Robotics X Lab (affiliation when this work was conducted) \\
  \footnotesize \textbf{Corresponding author:} Jingchen Li\texttt{<jingchenli\_njzy@163.com>}\\
}

\begin{document}

\twocolumn[ % Method A for two-column formatting
  \begin{@twocolumnfalse} % Method A for two-column formatting

\maketitle
\thispagestyle{empty}

\begin{abstract}
Service robots work with people and are highly expected to be lightweight for safety, 
agility and energy conservation. As a complex mechanical system, a robot consists of a 
large number of components and has various working configurations and load environments. 
An effective method for achieving system-level optimal robot design is a crucial 
requirement, but it poses significant challenges. In this study, we introduce a novel 
approach to optimize the distribution of strain energy, which can significantly improve 
the effectiveness of systematic optimization in a complex system. First, we present 
and demonstrate that the strain energy per unit mass should be uniformly distributed in an 
optimal lightweight mechanical system. Based on this criterion, the system-level problem can 
be decoupled and the design objective of each part can be assigned based on the strain 
energy. Then, each part can be optimally designed separately according to its specific 
circumstances by using different approaches, such as size optimization, topological 
optimization, and material optimization. In this way, the optimization is at the system 
level, while the computational complexity is at the part level. Weight reduction and improvements 
in mechanical properties can be obtained simultaneously. As an example, this method is applied 
to an arbitrarily designed robotic arm, and its effectiveness is further demonstrated in the 
cases of lightweight with stiffness improvement, considering multiple materials, multiple 
working conditions, and vibration performance. 
\end{abstract}
 % (optional)
\vspace{0.35cm}

  \end{@twocolumnfalse} % Method A for two-column formatting
] % Method A for two-column formatting

%\begin{multicols}{2} % Method B for two-column formatting (doesn't play well with line numbers), comment out if using method A

%%%%%%%%%%%%%%%  Main text   %%%%%%%%%%%%%%%
% \linenumbers

\section{Introduction}
\label{sec:intro}

Service robots are designed to interact with humans and operate in human environments, 
such as homes, hospitals, and public spaces. Therefore, they need to be lightweight and 
compact to move and manipulate dexterously and safely without causing harm to humans or 
damaging the environment~\cite{gonzalez2021service,alami2006safe}. Lightweight design also 
helps to reduce power consumption, which is important for service robots that are often 
battery-powered. Compared to other machines with limited functions, robots are capable of 
completing a wider range of tasks and operating in more complex work configurations and 
force environments. As a result, robots, especially robotic arms, legs or hands, usually 
require complex mechanical systems, and their lightweight design is often more intricate 
and challenging.  

Service robots are typically made up of mechanical structures, actuators, power systems, 
sensors and control systems, among which the mechanical structures and actuators contribute 
significantly to the weight~\cite{ding2021lightweight}. Actuators are essential components 
of robots and provide the necessary force and torque to move the joints and limbs, allowing 
the robot to perform a wide range of tasks. The primary approach to the lightweight design of 
actuators is to improve the power density. Modular or integrated designs are commonly used 
to achieve more compact actuators~\cite{albers2006upper,albu2007dlr,grimminger2020open}. 
Alternatively, many new drive modes are also used, such as elastic actuators~\cite{lenzi2018design,gabert2020compact}, 
rope drives~\cite{lens2013design,song2018development}, parallel linkage drives~\cite{tazaki2019parallel,reher2019dynamic,reher2016realizing,castillo2021robust}, 
and pneumatic drives~\cite{elena2018novel,ohta2018design,mori2018high}. In addition, 
appropriate selections of actuators based on dynamic simulation can also help to decrease 
the redundancy and obtain the systematic lightweight of actuators~\cite{zhou2011design,pettersson2009drive,ge2017optimization,padilla2018concurrent}.

The mechanical structures of a robot are responsible for providing support, stability, 
and mobility. They need to withstand the forces generated by robot movements and 
interactions with the environment. The strength and deformation of structures must be 
controlled at design time through necessary calculation and analysis, such as finite 
element analysis (FEA). However, stronger structures usually bring more weight. To achieve 
lightweight structures with adequate mechanical properties, topological optimization is an 
effective way and therefore has been widely used in various robots~\cite{albers2007methods,lohmeier2006leg,lohmeier2009humanoid,ye2014lightweight,albers2009integrated,sun2022larg,vazquez2019design,klemm2019ascento}. 
With the development of the additive manufacturing technology, structures from topological 
optimization can be manufactured more readily, which further improves the lightweight effect~\cite{junk2018topology,zhang2020lightweight}. 
Meanwhile, additive manufacturing also enables the application of lattice structures~\cite{hagenah2013modelling,folgheraiter2018design,jiang2023design,zong2024bionic}, 
which have better mechanical properties and lower weight than solid materials, and can 
bring significant weight reduction. Besides, some composite materials also have been used 
for the lightweight design of robots, such as the carbon fiber composite material~\cite{qi2004optimal,campeau2019kinova,kim2022application}. 
Composite materials usually have anisotropic mechanical properties, which make the 
manufacturing of complex structures difficult but provide the possibility of orientation 
optimization design of fibers and additive manufacturing~\cite{albers2009integrated}.

However, these lightweight methods of mechanical structures are mainly applied to individual 
parts. A robot is an assembly containing many parts, which collectively determine the mechanical 
properties of the system. A robot whose parts are separately designed to have maximum 
stiffness-to-mass ratios cannot have maximum stiffness at the system level~\cite{kim2016topology}. 
Therefore, systematic optimization is necessary. A direct idea is to implement the 
topological optimization of the whole assembly system synchronously. Since topological 
optimization needs iterative computations, and the computational complexity sharply increases 
with the density of meshes, a global topological optimization has much higher computational 
complexity. Another way is the size optimization, which is to optimize the design sizes, 
such as length, thickness and radius, to achieve the systematic optimization~\cite{yin2019hybrid,wang2022lightweight}. 
The global size optimization can also combine dynamic simulation and optimization of 
actuators~\cite{zhou2015new,yin2017unified,li2021integrated}. 
There are far fewer design variables in size optimization than in topological optimization, and 
thus the computational complexity is relatively lower. Obviously, size optimization is less 
effective than topological optimization. If the structures are very complex (as is usual in 
robotics), the parameterization of geometric sizes is also very difficult. Besides, it 
should be noted that the synchronous optimization of the whole assembly cannot use different 
optimization methods for different parts. For example, topological optimization is suitable 
for those large and complex parts, while using size or material optimization is more 
effective for some small and simple parts. For robot systems, due to the various tasks and 
working conditions, multi-condition optimization is usually necessary, which further 
increases the difficulty and computational complexity of the systematic optimization~\cite{wu2025timevarying}.

Therefore, decoupling the systematic problem into part-level optimization problems is necessary. 
To achieve this, the role of every part in the mechanical properties of the system should 
be recognized effectively to align the optimization objective of each part. The metamodel 
method~\cite{kim2016topology,wang2019optimal,hu2020multi,krischer2021decomposition,krischer2024distributed} was 
proposed from this viewpoint. Through repeatedly optimizing each part to obtain the 
stiffness-mass relationships, the proper set of mass constraints for part-level 
optimizations can be determined, and then the system-level problem can be decoupled. 
This method is systematically effective, but obtaining the nonlinear stiffness-mass 
relationships of the parts still requires substantial additional computational cost. To align the optimization 
objective of each part more directly and conveniently, we present a novel method based on 
the strain energy distribution optimization. Our motivation is that an optimal lightweight 
system means that each unit of mass should play the same role in the system. We use the scalar 
strain energy as the indicator and suppose that the strain energy per unit mass should be 
uniformly distributed among the whole mechanical system. Making the specific strain energy 
(i.e., strain energy per unit mass) of each part equal can help to assign the optimization objective of each part. Then each 
part can be optimized by various methods to approach the systematically assigned objective. 
Our systematic allocation and part-specific optimization method is demonstrated to be efficient 
in comparison with global optimization. As an application example, a robot arm 
is used to show how to define multiple load conditions, evaluate the stiffness, and optimize 
the weight under the expected performance by combining topological and material optimizations. 

\section{Strain Energy Distribution Optimization Method}
\label{sec:method}

An optimal robot system should have minimum weight and sufficient 
mechanical properties, including strength and stiffness to prevent 
failure, excessive deformation, and vibration. To achieve the minimum 
weight in theory, each unit of mass should play the same role in the system, 
i.e., the system should have neither redundancy nor weakness. This 
is a well-known principle and a matter of practical experience in structural design~\cite{galambos2008structural,ashby2012engineering}. 
Therefore, to optimize a system, the weak parts should be strengthened, 
and the redundant parts should be cut. Stress or strain is usually used 
to identify regions that are redundant or insufficient in a part. However, 
they are not applicable as indicators for quantitatively identifying 
whether a part is redundant or weak in a system. Firstly, strain and 
stress are both tensors, and their components in different directions 
play different roles. Secondly, they are only local properties and 
cannot be integrated to compare different parts. Besides, various 
materials have different constitutive relations, which should be 
considered when comparing parts made of different materials. 

The commonly used von Mises criterion is based on strain energy density to quantify the local 
stress state and evaluate structural strength. Simultaneously, 
the stiffness of a mechanical system can also be evaluated using strain 
energy, because it reflects the deformation under the load. 
\begin{equation}
    U_{tot}=\frac{1}{2}F_{ext}d=\iiint_{\Omega} u dV
    =\iiint_{\Omega} \frac{1}{2} 
    \boldsymbol{\varepsilon \sigma} dV.
\end{equation}
Here, $U_{tot}$ is the total strain energy of the system, 
$F_{ext}$ is the generalized external force, $d$ is the generalized 
deformation generated by the generalized external force, 
$\Omega$ is the whole volume domain of the mechanical system, 
$u$ is the strain energy density (per unit volume), 
and $\boldsymbol{\varepsilon}$ and $\boldsymbol{\sigma}$ are local 
strain and stress. Strain energy incorporates the material property 
that $\boldsymbol{\sigma} = \boldsymbol{E\varepsilon}$. 

Because strain energy is a scalar, incorporates material properties, and reflects structural strength 
and stiffness, it can be used to compare the contributions of different parts. 
To normalize for part mass, we use the specific strain energy, defined as the strain energy per unit mass, 
as the part-level indicator. Then, 
the optimization problem of a system becomes
\begin{align}
    \min & \ m_{tot}, \\
    s.t. & \ U_{tot} \le U_{tot}^{ref},    \\
         & \ u \le \frac{[u]}{SF}. 
\end{align}
Here, $m_{tot}$ is the total mass of the system, 
$U_{tot}^{ref}$ is the expected total strain energy 
(i.e., expected stiffness), $SF$ is the safety factor, 
and $[u]$ is the allowable strain energy density, 
obtained from the allowable stress of the corresponding material.
The two constraints are stiffness and strength conditions, 
respectively. Alternatively, the minimum stiffness can be the 
objective while the total weight can be the constraint, which is 
also a common form of optimization. 

Based on the viewpoint that each unit of mass should play the same role in 
the system, we hypothesize that the solution to the optimization problem 
is one in which the specific strain energy is uniformly 
distributed throughout the whole mechanical system. At the part level, 
the following condition should hold for all parts:
\begin{equation}
    \frac{U_i}{m_i} = \frac{U_{tot}^{ref}}{m_{tot}}
    \label{eq:hypothesis}
\end{equation}
Here, $U_i/m_i$ is the specific strain energy of part $i$ and should not be confused with 
$u$, the volumetric strain energy density defined above.

\subsection{Proof of the Strain Energy Distribution Criterion}

Next, we demonstrate that the hypothesis above (Equation~\ref{eq:hypothesis}) 
is valid. Although it has been tested in a previous study~\cite{kim2016topology} 
using two linked beams, here we further demonstrate it using 
more general methods. 

First, we consider an arbitrary number $n$ of rods in series. One end 
is fixed, and a force $F$ is applied to the other end. The rods do not need 
to be coaxial, so they can undergo tension, compression, torsion, 
bending, or combined deformation. Their cross-sections are all 
circular, and $r_i$ and $L_i$ are the cross-section radii and lengths, 
respectively. Each rod can be made of a different material, 
and $E_i$ and $\rho_i$ are the Young's modulus and material density, 
respectively. Based on the superposition principle, the total end 
deformation is the sum of the deformations of all rods induced by 
the end load $F$. The tension or compression deformation is
\begin{equation}
d_{i,ten}^{end} = \frac{FL_i}{E_i\pi r_i^2}.
\end{equation}
The torsional deformation (rotation angle) is
\begin{equation}
\varphi_{i,tor}=\frac{2 T_i L_i}{G_i \pi r_i^4} \propto \frac{F D_i L_i}{E_i r_i^4},
\end{equation}
where $T_i$ is the torque induced by the end force, and $D_i$ 
is the distance from the end of the $i^{\text{th}}$ rod to the system end. $G_i$ 
is the shear modulus. The rotation angle also induces an end deformation 
related to the distance $D_i$:
\begin{equation}
d_{i,tor}^{end} \propto \frac{F D_i^2 L_i}{E_i r_i^4}.
\end{equation}
The bending deformation includes a deflection and a rotation angle:
\begin{equation}
d_{i,bend,def}^{end}= \frac{4FL_i^3}{3E_i\pi r_i^4} 
\propto \frac{FL_i^3}{E_i r_i^4},
\end{equation}
\begin{equation}
\varphi_{i,bend}=\frac{4 F D_i L_i}{E_i \pi r_i^4}.
\end{equation}
The rotation angle also induces an end deformation related to the 
distance $D_i$:
\begin{equation}
d_{i,bend,rot}^{end} \propto \frac{F D_i^2 L_i}{E_i r_i^4}.
\end{equation}
All of these formulas are derived from mechanics-of-materials theory, which is based 
on the hypothesis that the rod length is much larger than the radius. 
The tension or compression deformation is much less than the torsional 
and bending deformations. Therefore, tension or compression 
deformation can be neglected, and the total end deformation is 
\begin{equation}
d_{end} = \sum_{i=1}^n (a\frac{FL_i^3}{E_i r_i^4} +b\frac{FD_i^2L_i}{E_i r_i^4}),
\end{equation}
where $a$ and $b$ are constants.

The strain energy and specific strain energy of rod $i$ are
\begin{equation}
V_i = \frac{a F^2 L_i^3+b F^2 D_i^2 L_i}{2E_i r_i^4}.
\end{equation}
\begin{equation}
v_i=\frac{V_i}{m_i} = \frac{a F^2 L_i^2+b F^2 D_i^2}{2\rho_i \pi  E_i r_i^6}.
\label{eq11} \end{equation}

\begin{figure}[t]
    \centering
    \includegraphics[width=3.3in]{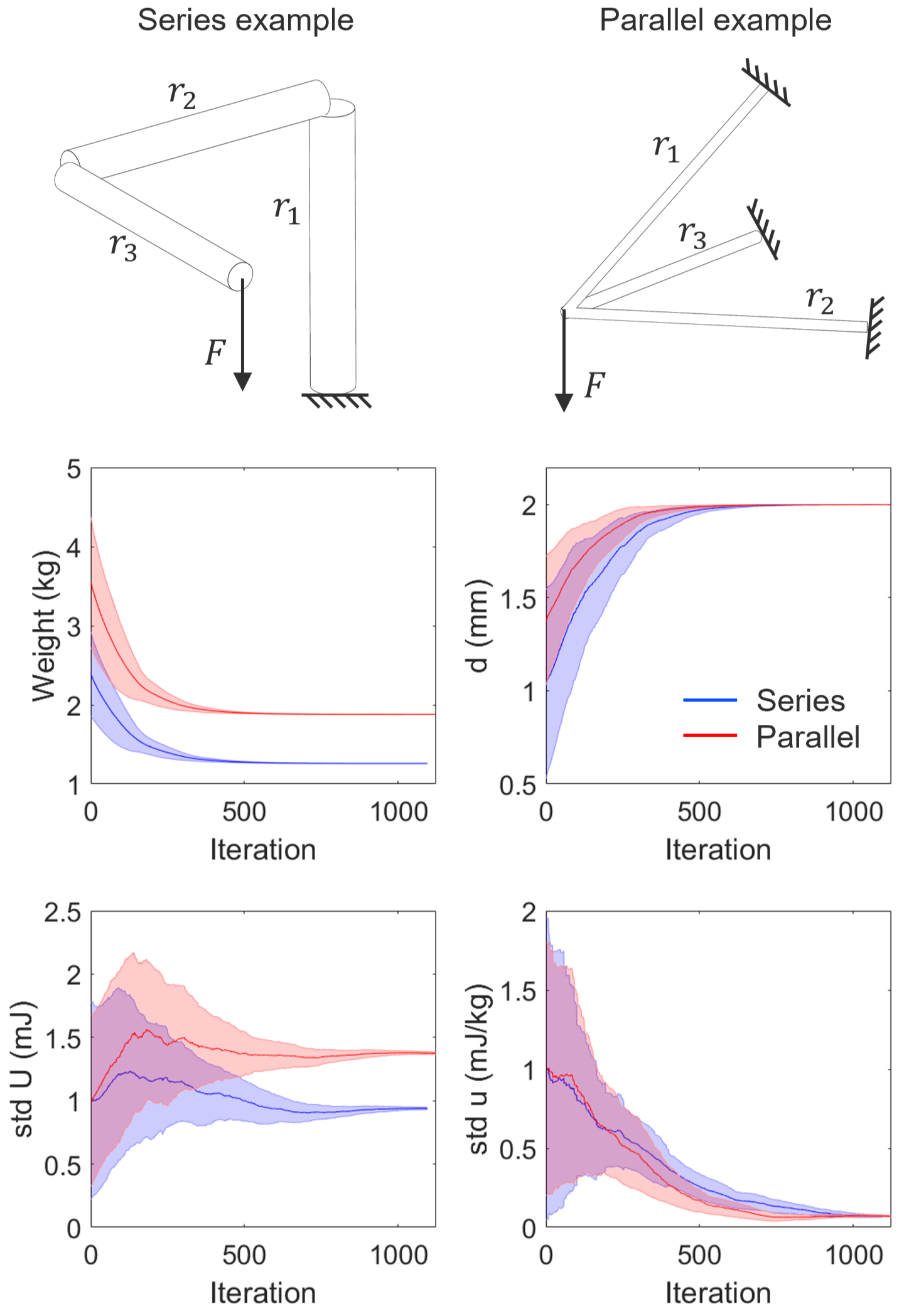}
    \caption{Two optimization examples using FEA. (1) Series example: three rods with circular cross sections are linked in series and are pairwise orthogonal. The root of the first rod is fixed, and a force $F$ is applied to the end of the last rod. (2) Parallel example: three rods with circular cross sections are linked in parallel and have $60^\circ$ angles between any two of them. The roots of the three rods are fixed, and a force $F$ is applied to the common end. For both examples, we solve an optimization problem with the objective of minimizing weight and the constraint that the end deformation is less than $2\,mm$. The bottom four panels show the total weight, end deformation, and the standard deviations (std) of the strain energies and specific strain energies of the three rods as the optimization progresses. }
    \label{fig1_beam_test}
\end{figure}

We define the optimization problem as
\begin{equation}
\min_{r_i} \ \ d_{end}=\sum_{i=1}^n (\frac{aFL_i^3+b F D_i^2 L_i}{E_i r_i^4} ),
\end{equation}
\begin{equation}
s.t. \ \ m_{tot}=\sum_{i=1}^n \rho_i \pi r_i^2 L_i=const.
\end{equation}
This is equivalent to minimizing the weight when the stiffness is 
maintained. The constraint condition can also be written as
\begin{equation}
r_n=\sqrt{\frac{m_{tot}-\sum_{i=1}^{n-1} \rho_i \pi r_i^2 L_i}{\rho_n \pi L_n}},
\end{equation}
and then
\begin{equation}
d_{end}=\sum_{i=1}^{n-1} (\frac{aFL_i^3+b F D_i^2 L_i}{E_i r_i^4} )+ \frac{aFL_n^3+b F D_n^2 L_n}{E_n r_n^4}.
\end{equation}
The optimal solution satisfies 
\begin{equation}
\frac{\partial d_{end}}{\partial r_i} =0.
\end{equation}
We obtain
\begin{equation}
\frac{aL_i^2+b D_i^2}{\rho_i  E_i r_i^6}=\frac{a L_n^2+b  D_n^2}{\rho_n E_n r_n^6}.
\label{eq17} \end{equation}
Obviously, by replacing $i$ with $i+1$, Equation~\ref{eq17} 
holds. Based on Equation~\ref{eq11}, we obtain 
\begin{equation}
\frac{v_{i+1}}{v_i}=1.
\end{equation}
Therefore, for a general series beam system, the optimal 
stiffness-to-weight ratio means that the specific strain energy 
of each part is equal. 

% 使优化后的重量和变形接近，选择：
% 串联系统，杆长200mm，F=100N
% 并联系统，杆长500mm，F=50000N

The above proof is based on mechanics of materials theory and is therefore 
valid only for a system of slender rods in series. Next, we use FEA 
to further verify it. As shown in Figure~\ref{fig1_beam_test}, we 
consider two cases: a series system and a parallel system. 

(1) In the series example, three rods with circular cross sections 
are linked in series and are pairwise orthogonal. The root of the first 
rod is fixed, and a force $F$ is applied to the end of the last rod. 

(2) In the parallel example, three rods with circular cross sections 
are linked in parallel and have 60-degree angles between any two of 
them. The roots of the three rods are fixed, and a force $F$ is 
applied to the common end. 

For these two examples, we keep all lengths fixed and optimize the 
three radii, $r_1$, $r_2$, and $r_3$, to achieve a lightweight design 
with a constraint on the end deformation, $d_{end}\le 2mm$. 
The three rods are made of structural steel, titanium alloy, and aluminum alloy, 
respectively. The optimization tool in ANSYS WORKBENCH is used, and the results 
are shown in Figure~\ref{fig1_beam_test}. Both cases show that the 
total weight decreases to convergence and the end deformation 
approaches the constraint as the optimization progresses. 
The specific strain energies of the three rods tend to become equal 
(their standard deviation approaches zero). In contrast, the strain 
energies of the three rods do not tend to become equal. These two cases 
further demonstrate that, in a mechanical system with the optimal 
stiffness-to-weight ratio, the strain energy-to-mass ratio of 
every part is equal; i.e., the specific strain energy is 
uniformly distributed. 

\subsection{Approach of Systematic Optimization}

We consider a robot system and specify a target stiffness. For 
example, when a load $F$ is applied to the robot, it has a corresponding 
deformation $d_{ref}$, and the target stiffness is 
$k_{ref}=F/d_{ref}$. The expected total strain energy is 
\begin{equation}
U_{tot}^{ref}=\frac{1}{2}F d_{ref}.
\label{eq19}
\end{equation}

A robot or mechanical system is made of many parts. Some parts 
contribute significantly to the weight and are mainly intended to provide 
supporting functions. These parts (labeled as $P$) should be mainly 
considered for lightweight optimization. However, some other parts 
not only need to resist the load but also have other functions, 
such as bearings, heat dissipation, and specialized connectors. These 
parts cannot be optimized solely based on achieving the expected 
system stiffness but need to consider other factors. We assume 
they have been optimally designed and no longer consider them 
(labeled as $N$). Through FEA, the strain energies of the initial 
system and each part can be obtained as 
\begin{equation}
U_{tot}=U_{rest}+\sum_{i=1}^{P} U_i=\sum_{j=1}^{N} U_j+\sum_{i=1}^{P} U_i,
\label{eq20}
\end{equation}
where $U_{rest}$ is the strain energy of those parts that will not 
be optimized and will remain unchanged during optimization. Based 
on the strain energy distribution criterion, each part should have 
the expected strain energy 
\begin{equation}
U_i^{ref}=\frac{U_{tot}^{ref}-U_{rest}}{\sum_{i=1}^P m_i^{opt}} m_i^{opt}.
\label{eq21}
\end{equation}
Therefore, each part has an optimization objective 
$U_i\rightarrow U_i^{ref}$. Here, $m_i^{opt}$ is the mass of the 
$i^{\text{th}}$ part after optimization and is unknown before optimization. 
To address this issue, we propose three methods as follows.

(1) Using the mass before optimization, $m_i^{ori}$, instead, 
\begin{equation}
U_i^{ref}=\frac{U_{tot}^{ref}-U_{rest}}{\sum_{i=1}^P m_i^{ori}} m_i^{ori}.
\label{eq22}
\end{equation}
This method is the most convenient but has obvious errors. Because 
the mass of each part changes after optimization, the final 
specific strain energy will not be uniform. However, this 
approximation is somewhat reasonable. An experienced designer can 
provide a relatively reasonable initial design, in which the part 
subjected to a larger load has a larger size and mass. A relatively 
reasonable initial mass distribution may make this method effective.

\begin{figure*}[t]
    \centering
    \includegraphics[width=6.8in]{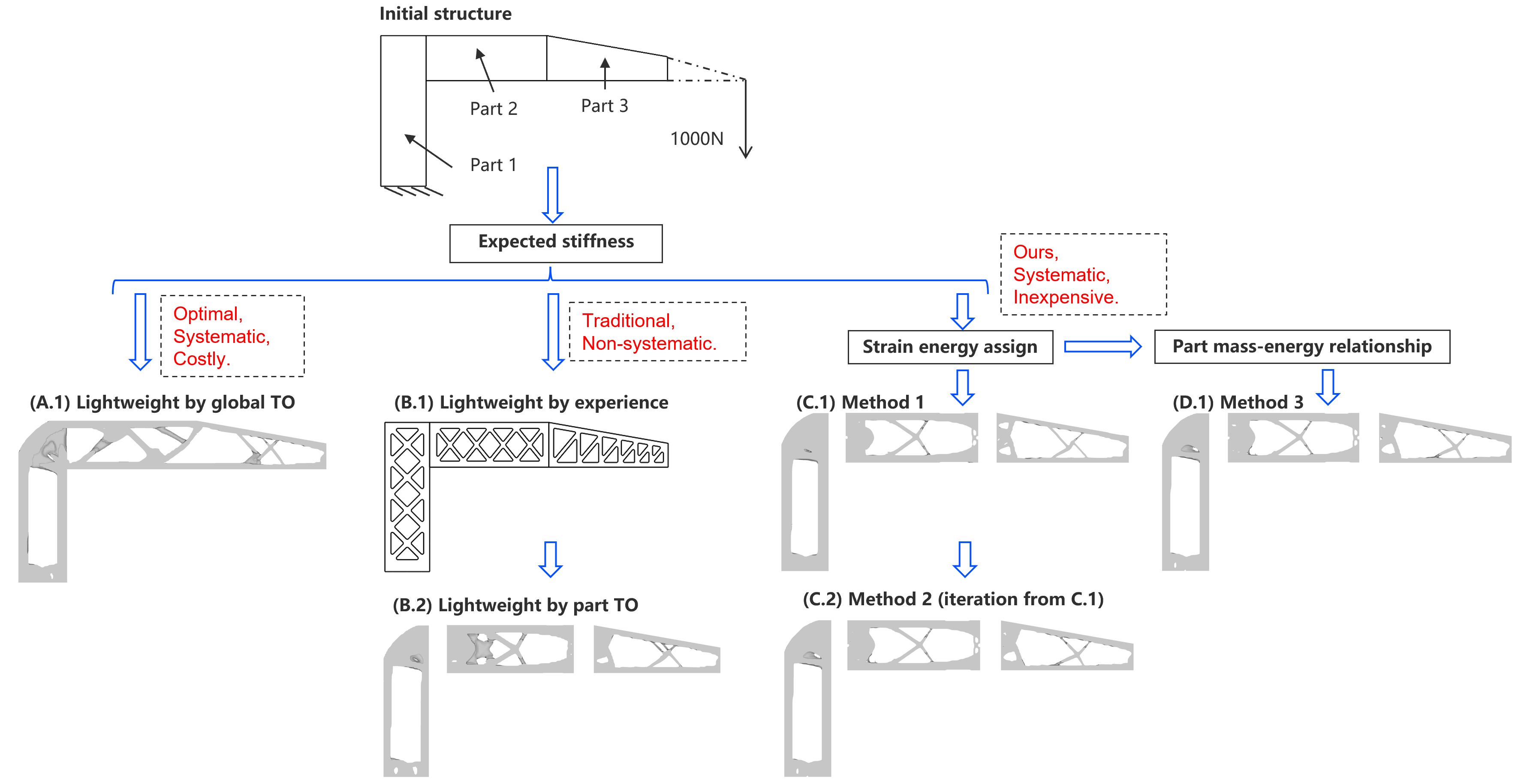}
    \caption{A simple example for testing the optimization methods. 
    The top panel shows the model setup. 
    After specifying a target stiffness, 
we compare global optimization, an experience-based method, and 
    our methods for systematic lightweight design. }
    \label{fig2_method_test}
\end{figure*}

\begin{figure}[t]
    \centering
    \includegraphics[width=3.2in]{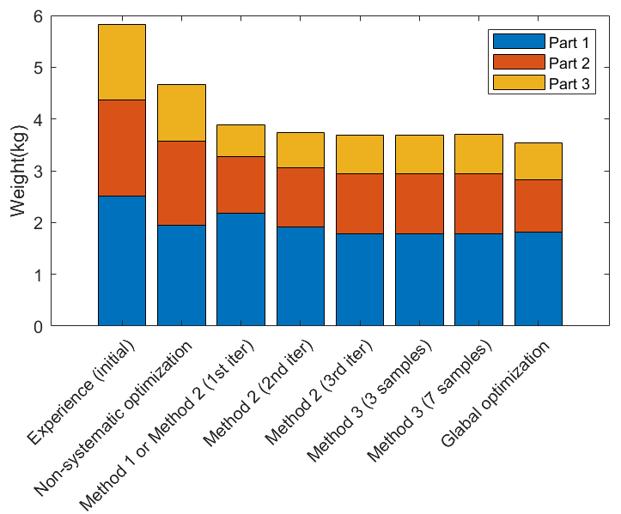}
    \caption{Comparison of lightweight results obtained using 
    the different methods shown in Figure~\ref{fig2_method_test}.
    The weights are sorted from largest to smallest.}
    \label{fig3_compare}
\end{figure}

(2) Using iteration: the first iteration uses the original mass, 
$m_i^{opt,(0)}=m_i^{ori}$, and then optimization of all 
parts is performed based on the previous optimization:
\begin{equation}
U_i^{ref,(n+1)}=\frac{U_{tot}^{ref}-U_{rest}}{\sum_{i=1}^P m_i^{opt,(n)}} m_i^{opt,(n)}.
\label{eq23}
\end{equation}
This method incurs additional computational cost but may be 
acceptable if convergence is rapid. We assume that there are two parts, 
and their masses after the $i^{\text{th}}$ iteration are $m_1^{(i)}$ and 
$m_2^{(i)}$. A larger mass usually means greater stiffness, and the 
strain energy is inversely proportional to the mass. We can assume 
$U\propto 1/m^{r}$~\cite{kim2016topology}. In the next iteration,
\begin{equation}
\frac{U_1^{ref,(i+1)}}{U_2^{ref,(i+1)}} = 
\frac{m_1^{(i)}}{m_2^{(i)}} \propto 
[\frac{m_2^{(i+1)}}{m_1^{(i+1)}}]^r.
\end{equation}
The mass ratio may oscillate because the new mass ratio is 
inverse to the old one at each step. Obviously, if $r > 1$, the iteration 
converges. In fact, $r$ may be larger than 1~\cite{kim2016topology}. 
To further ensure convergence and increase the convergence rate, 
we use the average mass of previous iterations for the next 
iteration:
\begin{equation}
U_i^{ref,(n+1)}=\frac{U_{tot}^{ref}-U_{rest}}{\sum_{i=1}^P m_i^{ave,(n)}} m_i^{ave,(n)},
\label{eq25}
\end{equation}
\begin{equation}
m_i^{ave,(n)}= \frac{1}{n} \sum_{k=1}^n m_i^{opt,(k)}.
\end{equation}

(3) Using the metamodel method~\cite{kim2016topology} to calculate 
the expected strain energy. Through repeated optimization of one 
part, the relationship between strain energy and the corresponding 
minimum mass can be obtained as 
\begin{equation}
    U_i^{ref}=fun(m_i^{opt}).
\end{equation}
By substituting this into Equation~\ref{eq21}, the expected strain energy 
$U_i^{ref}$ can be solved. This method is obviously effective, but 
additional calculations are also needed. To obtain the nonlinear relationship, 
optimization of each part needs to be performed at least three times, 
and then the expected strain energies are obtained to perform the 
final optimization. Therefore, at least four calculations for 
each part are needed.

Comparing the three methods above, the first one has the lowest 
computational cost but the largest error. The latter two methods 
both require iterations. The second method is a global iteration, and 
each step can yield a better final result. Perhaps the convergence 
is fast and the result can be acceptable after a few steps, whereas 
the third method is an individual iteration. After the nonlinear 
mass-stiffness relationship is obtained, the objective can be 
calculated, and then one additional optimization is needed. Therefore, 
the second method seems convenient and may have the lowest 
computational cost. 

\section{Test and Compare the Methods}

We use a simple example to compare the methods above. As shown 
in Figure~\ref{fig2_method_test}, a remote force is applied to the end of 
a quasi-two-dimensional structure made of three parts. All parts are 
made of aluminum alloy, and the initial weight is about $10 kg$. 

Since the structure is simple, we can directly use global topological 
optimization for all three parts simultaneously to obtain a system-level 
optimal design (Figure~\ref{fig2_method_test} A.1). The optimization 
objective is minimum weight, and the constraint is that the end 
deformation is less than $4.5 mm$. For simplicity, we directly compare 
the weight and deformation of the element-density results from topology 
optimization but do not redesign the parts in this section. The 
optimization result has a total weight of $3.54 kg$ and an end 
deformation of $4.5 mm$. This result can serve as a reference for comparing 
different methods. 

In fact, global topological optimization is not feasible in real 
robot systems. Next, we test and compare the methods proposed above 
to verify whether part-level optimization can achieve a system-level effect.

\begin{figure}[t]
    \centering
    \includegraphics[width=2.8in]{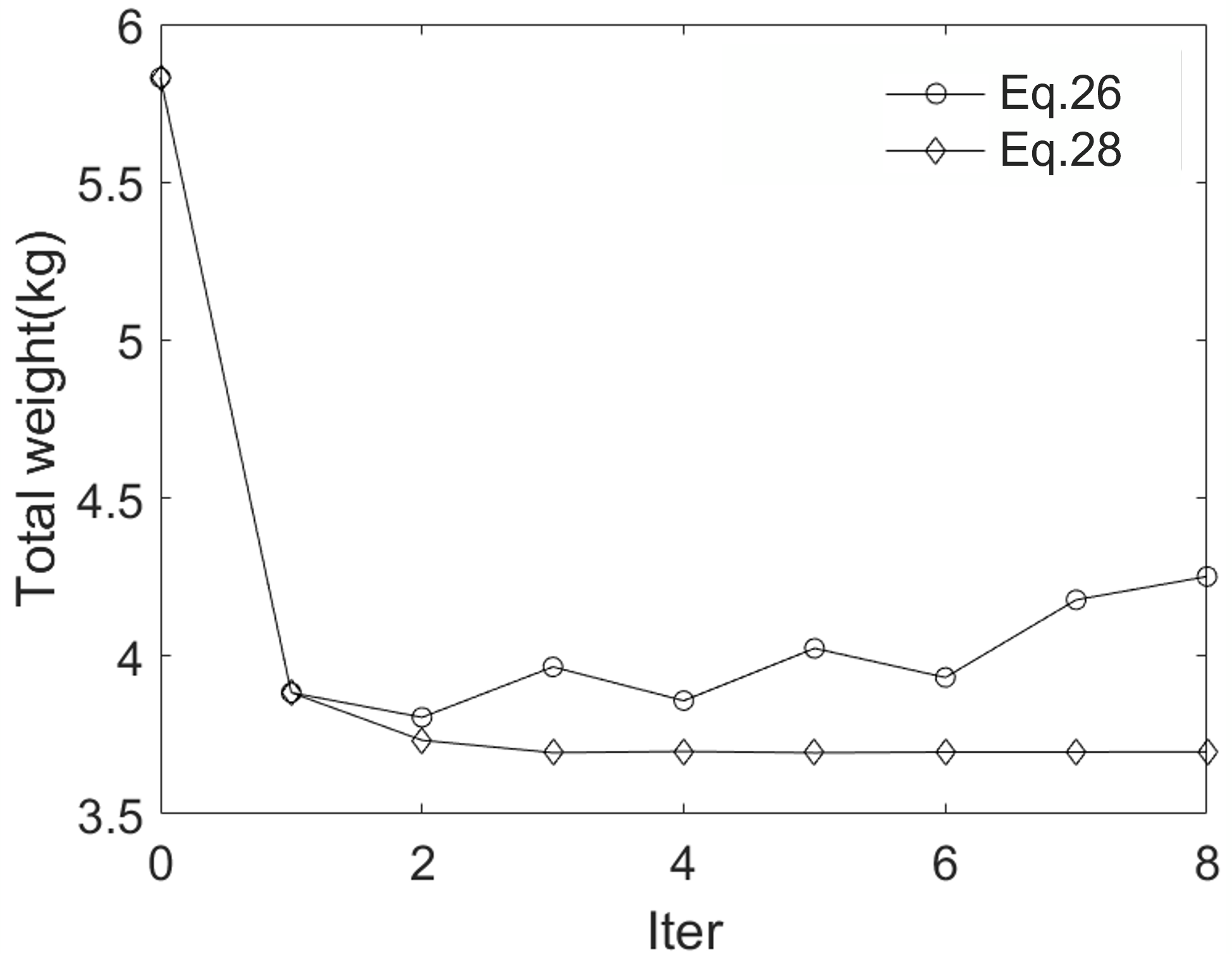}
    \caption{The total weight of the example system 
    (Figure~\ref{fig2_method_test}) converges with the iteration 
    when using Method 2.}
    \label{fig4_convergence}
\end{figure}

\begin{figure}[t]
    \centering
    \includegraphics[width=3.2in]{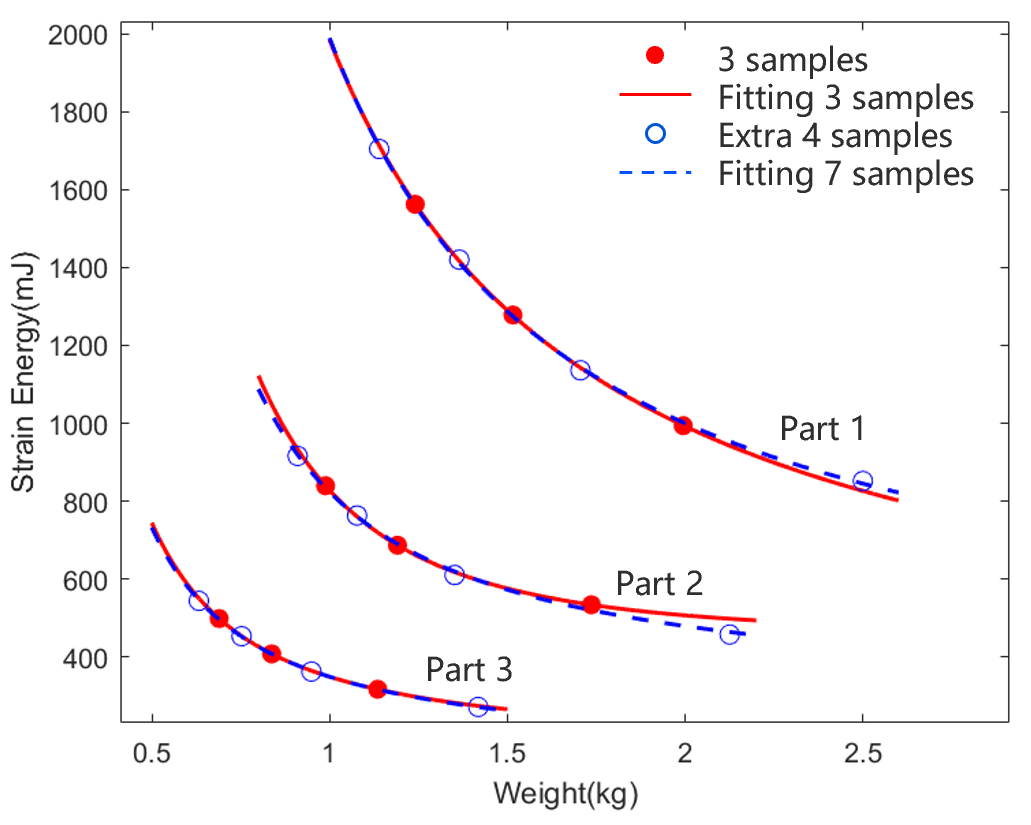}
    \caption{The relationship between the optimized weight and the 
    strain energy of each part of the example system 
    (Figure~\ref{fig2_method_test}) when using Method 3.}
    \label{fig5_meta_model}
\end{figure}

\subsection{Traditional Lightweight Methods}

First, we use traditional lightweight methods. Some material 
can be removed based on experience, and the mechanical 
properties can be analyzed by FEA. Iterative design and 
simulation can ensure the mechanical properties. 
Figure~\ref{fig2_method_test} B.1 gives an example. The end 
deformation of this structure reaches $4.5 mm$. Obviously, this 
method is not optimal, and the total weight is reduced to $5.83 kg$. 
However, it is difficult to further reduce the weight intuitively 
without topological optimization. 

Due to the high computational cost of systematic topological 
optimization, part-level topological optimization is usually used. 
A common practice is to maximize the 
stiffness (minimize the compliance) with the 
constraint of reducing the mass of every part by the same ratio. This method cannot 
ensure that the mechanical properties remain within a specified range. Another 
common method is to minimize weight with 
the mechanical properties as constraints. The ratio of the system 
stiffness before optimization to the expected system stiffness is 
equally applied to each part as the constraint for topological 
optimization. As an example (B.2), the lightweight design using 
this method can ensure that the deformation is less than $4.5 mm$ and 
has a total weight of $4.67 kg$, which is lighter than the 
experience-based design but appreciably heavier than the global 
optimal design (Figure~\ref{fig3_compare}). Obviously, these 
part-level optimizations cannot achieve systematic optimization. 
Although every part is separately designed to have the maximum 
stiffness-to-mass ratio, the system cannot have the maximum 
stiffness-to-weight ratio. The reason is that every part is treated 
equally in the system, which implies the assumption that every part 
plays the same role in the system before optimization. The initial 
design hardly has that feature. If one part is too strong and 
another is too weak, the topological optimization result of the 
corresponding part remains unchanged, although the system stiffness constraint can 
be satisfied.

\subsection{Strain Energy Distribution Optimization Method}

Unlike the traditional methods above, we use strain energy 
distribution to identify the different roles of every part in the 
system and optimally assign the optimization conditions of each part; 
thereby, systematic optimization can be achieved by part-level 
optimization. First, the initial design is simulated by FEA, and 
the strain energy of each part can be obtained. Then, we use 
method 1 (Equation~\ref{eq22}) to calculate the expected strain 
energy of each part. Topological optimization is then used for 
each part to approach the objective. The optimal results are shown 
in Figure~\ref{fig2_method_test} C.1. We obtain a lightweight 
design with a total weight of $3.88 kg$, which is relatively close 
to the global optimization result (Figure~\ref{fig3_compare}).

The results of method 1 are also the first iterative step of 
method 2. We then continue the iteration. We have proposed two 
iteration methods (Equations~\ref{eq23} and~\ref{eq25}) and 
guessed that the latter may have better convergence. As shown in 
Figure~\ref{fig4_convergence}, when the mass from the last 
optimization is used to perform the next iteration (Equation~\ref{eq23}), 
the iteration diverges. When the average mass of previous iterations 
is used for the next iteration (Equation~\ref{eq25}), it converges quickly. 
After each step, a lighter design is obtained, and by the 
third step, the iteration has already converged. Using method 2, we 
obtain a lightweight design with a total weight of $3.69 kg$ 
(Figure~\ref{fig2_method_test} C.2), which is closer to the global 
optimization result (Figure~\ref{fig3_compare}).

Finally, we use method 3. To fit the nonlinear relationship 
between mass and stiffness, we first generate three samples 
by performing topological optimization of each part with different 
stiffness constraints. We use the following metamodel~\cite{kim2016topology}:
\begin{equation}
    U_i^{ref}= \sum_{j=1}^3 b_{ij} m_i^{-q_j}.
\label{eq28}
\end{equation}
Here, $b_{ij}\ge0$ and $q_j \ge 0$ $(j=1,2,3)$ are obtained by 
minimizing the sum of squared differences between the metamodel 
and the topology optimization results at the sample points 
(Figure~\ref{fig5_meta_model}). For comparison, we further 
generate four additional samples to fit the metamodel. 
The fitting curves are highly similar (Figure~\ref{fig5_meta_model}). 
Then, Equation~\ref{eq28} is substituted into Equation~\ref{eq21} 
to solve for the expected strain energy $U_i^{ref}$. The optimal 
results are shown in Figure~\ref{fig2_method_test} D.1. The weight 
is $3.69 kg$, approximating the result of method 2 
(Figure~\ref{fig3_compare}). The optimal topological structures of 
C.2 and D.1 are also almost the same (Figure~\ref{fig2_method_test}), 
which indicates that methods 2 and 3 are equivalent. 

There is still a small gap between our methods and the global 
optimization method. By observing the topological differences 
(Figures~\ref{fig2_method_test} A.1, C.2, and D.1), we can find that 
the connection regions of each part are stronger in the results of 
our methods. This is because, when optimizing one part, the other 
parts are not included in the FEA model to avoid system-level 
computational cost. $REB3$ elements are used to equivalently 
simulate the boundary conditions, connections to other parts, and 
the remote force. $REB3$ elements make the connection interface 
weaker than direct connections, which causes the stiffness to 
be underestimated and thus results in a larger mass after 
optimization.

\begin{figure}[t]
    \centering
    \includegraphics[width=3.4in]{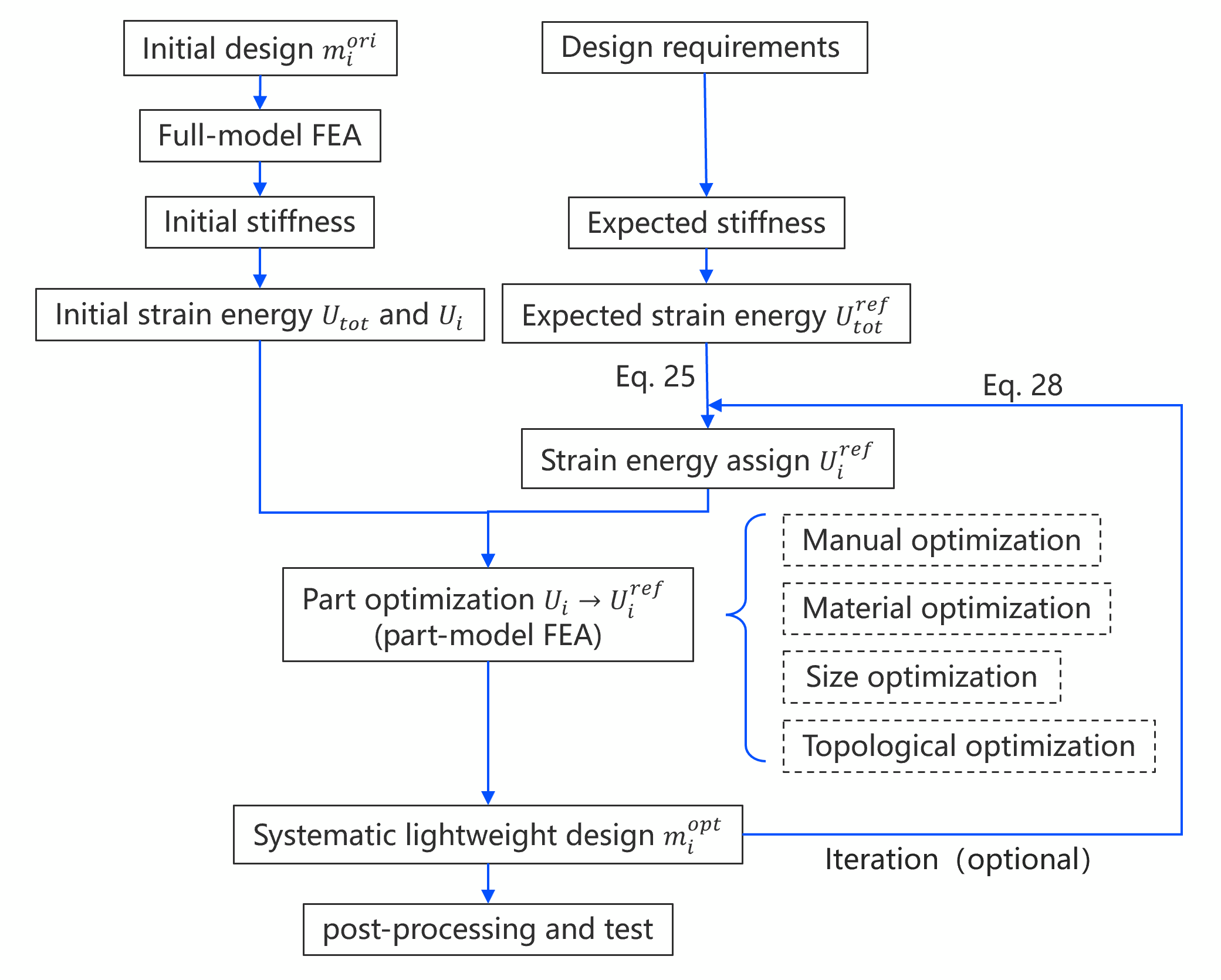}
    \caption{The basic steps of the proposed strain-energy-based optimization method.}
    \label{fig6_flow_chart}
\end{figure}

Comparing all the methods above (Figures~\ref{fig2_method_test} 
and~\ref{fig3_compare}), the global optimization method can provide 
the optimal design. Our methods can provide designs that approach 
the optimal design. It is verified that our methods achieve systematic 
optimization and are better than the non-systematic methods. 
Methods 2 and 3 are equivalent and can provide better results than 
method 1, but method 3 is more costly. Methods 1 and 2 are 
recommended for different trade-offs between computational cost and 
error. As shown in Figure~\ref{fig6_flow_chart}, the basic steps of 
the strain-energy-based optimization methods are presented. 

\begin{figure*}[t]
    \centering
    \includegraphics[width=6.8in]{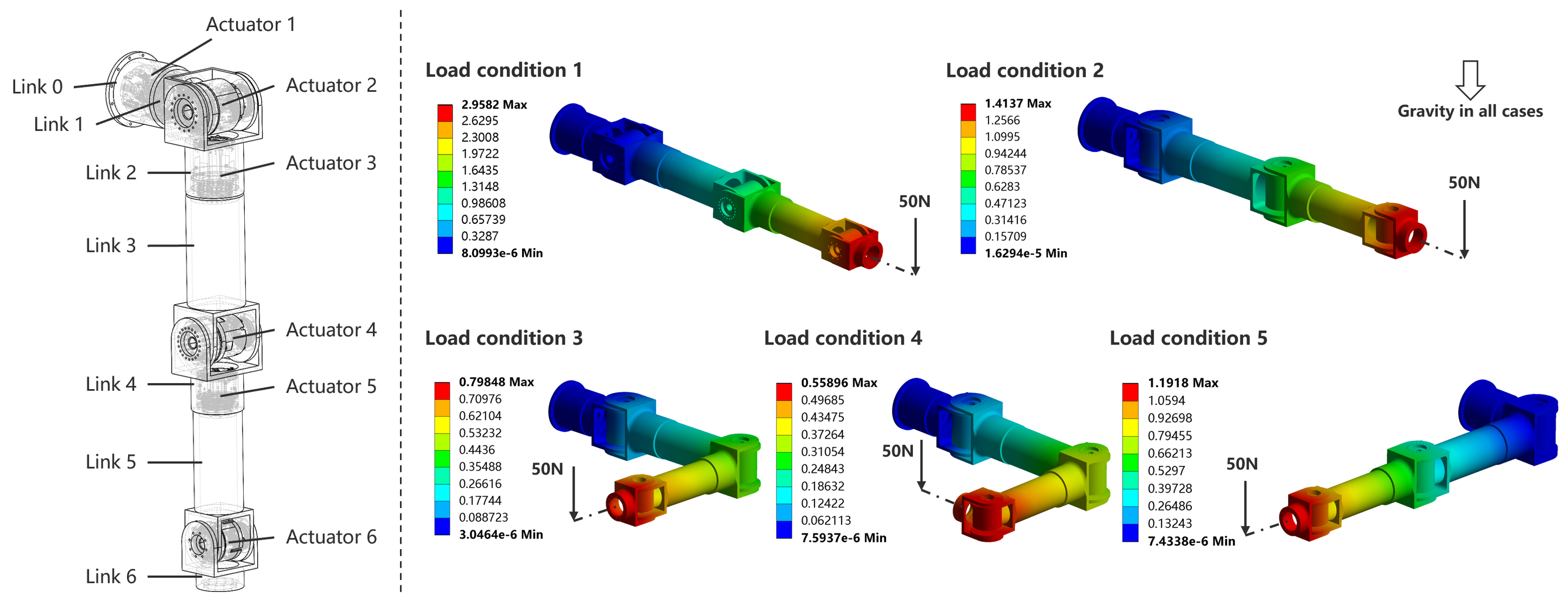}
    \caption{(Left) A robotic arm with a typical design. 
    (Right) The 5 horizontal poses contain all the worst loads on each link, and the deformation results are shown from FEA. }
    \label{fig7_arm_sim0}
\end{figure*}

\begin{table*}[htbp]
    \centering
    \caption{Equivalent load on each link in each condition.}
    \label{tab1}
    \begin{threeparttable}
        \begin{tabular}{cccccc}
        \hline
        & Cd 1 & Cd 2 & Cd 3 & Cd 4 & Cd 5 \\
        \hline
        Link 0 & 
        $\bar{F}_3$, $\bar{T}_1$ & 
        $\bar{F}_3$, $\bar{T}_1$ &
        $\bar{F}_3$, $T_1$, $T_2$ & 
        $\bar{F}_3$, $T_1$, $T_2$ & 
        $\bar{F}_3$, $T_1$, $\bar{T}_2$ \\
        %\hline
        Link 1 & 
        $\bar{F}_3$, $\bar{T}_1$ & 
        $\bar{F}_1$, $\bar{T}_3$ & 
        $\bar{F}_1$, $T_2$, $T_3$ & 
        $\bar{F}_1$, $T_2$, $T_3$ & 
        $\bar{F}_1$, $\bar{T}_2$ \\
        %\hline
        Link 2 &
        $\bar{F}_3$, $\bar{T}_1$  &
        $\bar{F}_1$, $\bar{T}_3$   &
        $\bar{F}_1$, $\bar{T}_2$, $T_3$    &
        $\bar{F}_1$, $T_2$, $T_3$     &
        $\bar{F}_1$, $\bar{T}_3$ \\
        %\hline
        Link 3 &
        $\bar{F}_3$, $\bar{T}_1$  &
        $\bar{F}_1$, $\bar{T}_3$   &
        $\bar{F}_1$, $\bar{T}_2$    &
        $\bar{F}_1$, $T_2$, $T_3$     &
        $\bar{F}_1$, $\bar{T}_3$      \\
        %\hline
        Link 4 &
        $\bar{F}_3$, $\bar{T}_1$  &
        $\bar{F}_1$, $\bar{T}_3$   &
        $\bar{F}_1$, $\bar{T}_3$    &
        $\bar{F}_1$, $\bar{T}_2$, $T_3$     &
        $\bar{F}_1$, $\bar{T}_3$      \\
        %\hline
        Link 5 &
        $\bar{F}_3$, $\bar{T}_1$  &
        $\bar{F}_1$, $\bar{T}_3$   &
        $\bar{F}_1$, $\bar{T}_3$    &
        $\bar{F}_1$, $\bar{T}_2$     &
        $\bar{F}_1$, $\bar{T}_3$      \\
        %\hline
        Link 6 &
        $\bar{F}_3$, $\bar{T}_1$  &
        $\bar{F}_1$, $\bar{T}_3$   &
        $\bar{F}_1$, $\bar{T}_3$    &
        $\bar{F}_1$, $\bar{T}_3$     &
        $\bar{F}_1$, $\bar{T}_3$      \\
        \hline
        \end{tabular}
        \begin{tablenotes}
            \item[*] $F$ and $T$ denote force and torque on the link respectively generated by the end load. 
            The bar means it reaches the static maximal value. 
            Indices 1--3 denote the directional components relative to the local frame of the link. 
            For each link, the end load always has no component in some directions at horizontal poses in static cases.
            Cd is the abbreviation of Load condition.
        \end{tablenotes}
    \end{threeparttable}
\end{table*}

\begin{table*}[htbp]
    \centering
    \caption{Materials used in the optimization.}
    \label{tab2}
    \begin{tabular}{ccccc}
      \hline
      Material & $\rho$ ($kg/m^3$) & $E$ ($GPa$) & Poisson ratio & $E/\rho$ \\
      \hline
      Structural Steel & 7850 & 200 & 0.3 & 0.025 \\
      Titanium Alloy & 4500 & 110 & 0.36 & 0.024 \\
      Aluminium Alloy & 2770 & 71 & 0.33 & 0.026 \\
      Magnesium Alloy & 1800 & 45 & 0.35 & 0.025 \\
      \hline
    \end{tabular}
\end{table*}

\begin{table*}[htbp]
    \centering
\caption{Comparison of strain energies for the initial and target designs (the $2\,mm$ case as an example)}
    \label{tab3}
    \begin{tabular}{cc|cccc|c|cccc}
        \hline
         & & \multicolumn{4}{c|}{Initial Design} &  & \multicolumn{4}{c}{After Material Optimization} \\
        Cd & Part & $m_i^{ori}(kg)$ & $U_i(mJ)$ & $U_i^{ref}(mJ)$ & $U_i^{ref}$/$U_i$ & Comment & $m_i^{ori}(kg)$ & $U_i(mJ)$ & $U_i^{ref}(mJ)$ & $U_i^{ref}$/$U_i$\\ 
        \hline
        1 & Link 0 & 0.815 & 8.712 & 11.585 & 1.33 &  & 1.324 & 5.623 & 12.232 & 2.175 \\
        1 & Link 1 & 0.774 & 45.821 & 11.004 & 0.24 & Weak point & 2.18 & 16.266 & 20.139 & 1.238 \\
        1 & Link 2 & 1.107 & 50.473 & 15.726 & 0.312 & Weak point & 3.116 & 17.918 & 28.78 & 1.606 \\
        1 & Link 3 & 1.682 & 13.595 & 23.907 & 1.759 &  & 1.682 & 13.595 & 15.538 & 1.143 \\
        1 & Link 4 & 0.861 & 8.593 & 12.236 & 1.424 &  & 1.399 & 5.547 & 12.919 & 2.329 \\
        1 & Link 5 & 1.354 & 2.084 & 19.241 & 9.233 &  & 0.88 & 3.288 & 8.126 & 2.471 \\
        1 & Link 6 & 0.418 & 0.857 & 5.933 & 6.921 &  & 0.271 & 1.353 & 2.506 & 1.853 \\
        \hline
        2 & Link 0 & 0.815 & 8.71 & 12.808 & 1.47 &  & 1.324 & 5.622 & 13.523 & 2.405 \\
        2 & Link 1 & 0.774 & 24.486 & 12.165 & 0.497 & Weak point & 2.18 & 8.692 & 22.264 & 2.561 \\
        2 & Link 2 & 1.107 & 10.178 & 17.385 & 1.708 &  & 3.116 & 3.613 & 31.817 & 8.805 \\
        2 & Link 3 & 1.682 & 10.683 & 26.43 & 2.474 &  & 1.682 & 10.683 & 17.177 & 1.608 \\
        2 & Link 4 & 0.861 & 2.33 & 13.527 & 5.806 &  & 1.399 & 1.504 & 14.282 & 9.498 \\
        2 & Link 5 & 1.354 & 2.029 & 21.271 & 10.483 &  & 0.88 & 3.201 & 8.983 & 2.806 \\
        2 & Link 6 & 0.418 & 0.317 & 6.56 & 20.696 &  & 0.271 & 0.5 & 2.77 & 5.54 \\
        \hline
        3 & Link 0 & 0.815 & 4.371 & 14.207 & 3.25 &  & 1.324 & 2.822 & 15.0 & 5.316 \\
        3 & Link 1 & 0.774 & 14.224 & 13.495 & 0.949 & Weak point & 2.18 & 5.049 & 24.697 & 4.891 \\
        3 & Link 2 & 1.107 & 6.562 & 19.285 & 2.939 &  & 3.116 & 2.329 & 35.294 & 15.152 \\
        3 & Link 3 & 1.682 & 4.801 & 29.318 & 6.107 &  & 1.682 & 4.801 & 19.054 & 3.969 \\
        3 & Link 4 & 0.861 & 1.503 & 15.005 & 9.986 &  & 1.399 & 0.97 & 15.843 & 16.335 \\
        3 & Link 5 & 1.354 & 2.029 & 23.595 & 11.628 &  & 0.88 & 3.201 & 9.965 & 3.113 \\
        3 & Link 6 & 0.418 & 0.317 & 7.276 & 22.957 &  & 0.271 & 0.5 & 3.073 & 6.145 \\
        \hline
        4 & Link 0 & 0.815 & 3.151 & 19.066 & 6.051 &  & 1.324 & 2.034 & 20.13 & 9.898 \\
        4 & Link 1 & 0.774 & 10.346 & 18.109 & 1.75 &  & 2.18 & 3.673 & 33.142 & 9.023 \\
        4 & Link 2 & 1.107 & 4.14 & 25.88 & 6.251 &  & 3.116 & 1.47 & 47.363 & 32.226 \\
        4 & Link 3 & 1.682 & 2.229 & 39.344 & 17.652 &  & 1.682 & 2.229 & 25.57 & 11.472 \\
        4 & Link 4 & 0.861 & 0.848 & 20.136 & 23.748 &  & 1.399 & 0.547 & 21.26 & 38.847 \\
        4 & Link 5 & 1.354 & 1.032 & 31.664 & 30.672 &  & 0.88 & 1.629 & 13.373 & 8.21 \\
        4 & Link 6 & 0.418 & 0.126 & 9.765 & 77.225 &  & 0.271 & 0.199 & 4.124 & 20.671 \\
        \hline
        5 & Link 0 & 0.815 & 3.002 & 11.657 & 3.882 &  & 1.324 & 1.938 & 12.308 & 6.351 \\
        5 & Link 1 & 0.774 & 18.985 & 11.072 & 0.583 & Weak point & 2.18 & 6.74 & 20.263 & 3.007 \\
        5 & Link 2 & 1.107 & 8.02 & 15.823 & 1.973 &  & 3.116 & 2.847 & 28.958 & 10.171 \\
        5 & Link 3 & 1.682 & 10.683 & 24.055 & 2.252 &  & 1.682 & 10.683 & 15.634 & 1.463 \\
        5 & Link 4 & 0.861 & 2.33 & 12.311 & 5.285 &  & 1.399 & 1.504 & 12.999 & 8.644 \\
        5 & Link 5 & 1.354 & 2.029 & 19.359 & 9.541 &  & 0.88 & 3.201 & 8.176 & 2.554 \\
        5 & Link 6 & 0.418 & 0.317 & 5.97 & 18.836 &  & 0.271 & 0.5 & 2.521 & 5.042 \\
        \hline
    \end{tabular}
\end{table*}

\begin{figure*}[t]
    \centering
    \includegraphics[width=6.8in]{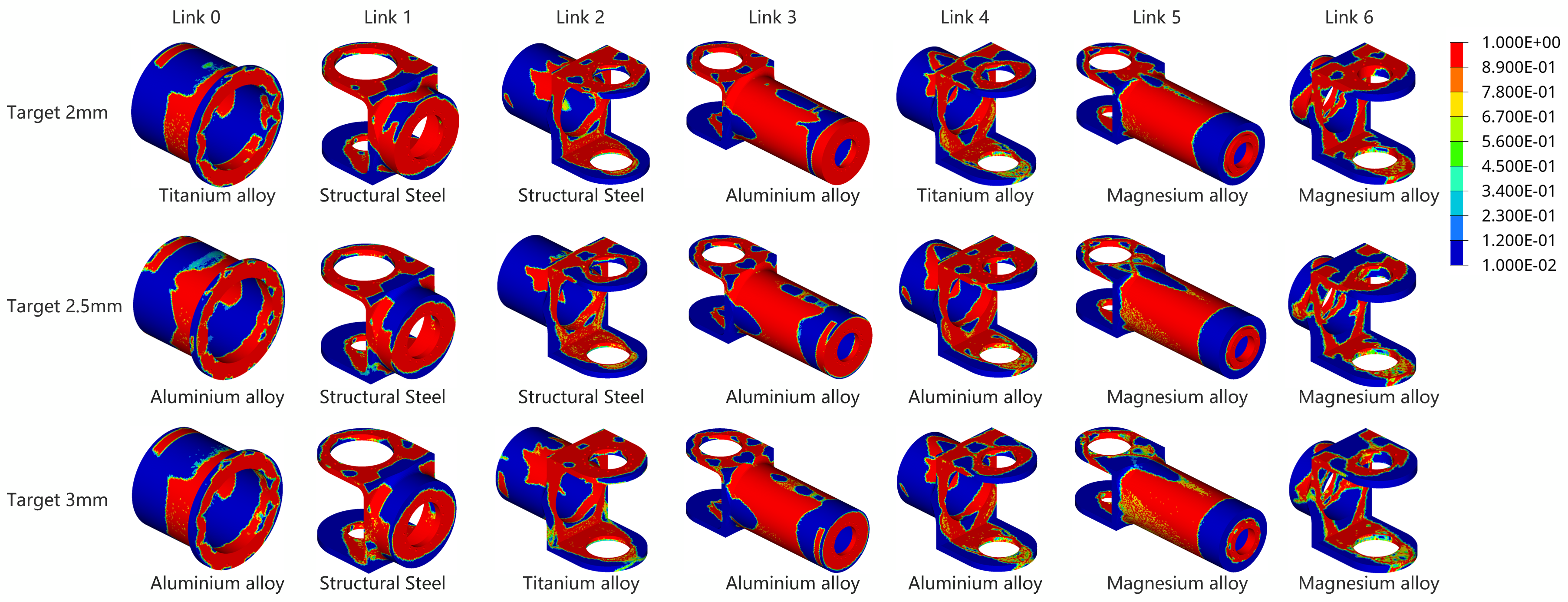}
    \caption{Density distributions of elements obtained by topological optimization.}
    \label{fig8_arm_TO}
\end{figure*}

\begin{table*}[htbp]
    \centering
    \caption{Optimization Results Summary}
    \label{tab4}
    \begin{tabular}{ccccc}
        \hline
        & Initial design & 2mm target & 2.5mm target & 3mm target \\
        \hline
        Total weight (kg) & 15.196 & 12.346 & 11.378 & 10.904 \\
        Structural weight (kg) & 7.116 & 4.266 & 3.298 & 2.824 \\
        Structural weight ratio & 46.8\% & 34.6\% & 29.0\% & 25.9\% \\
        \hline
        Max deformation of Cd 1 (mm) & 2.958 & 1.759 & 2.157 & 2.573  \\
        Max deformation of Cd 2 (mm) & 1.414 & 1.436 & 1.644 & 1.969  \\
        Max deformation of Cd 3 (mm) & 0.798 & 1.022 & 1.226 & 1.440  \\
        Max deformation of Cd 4 (mm) & 0.559 & 0.659 & 0.794 & 0.935  \\
        Max deformation of Cd 5 (mm) & 1.192 & 1.236 & 1.497 & 1.701  \\
        \hline 
    \end{tabular}
\end{table*}

\begin{figure}[t]
    \centering
    \includegraphics[width=3.3in]{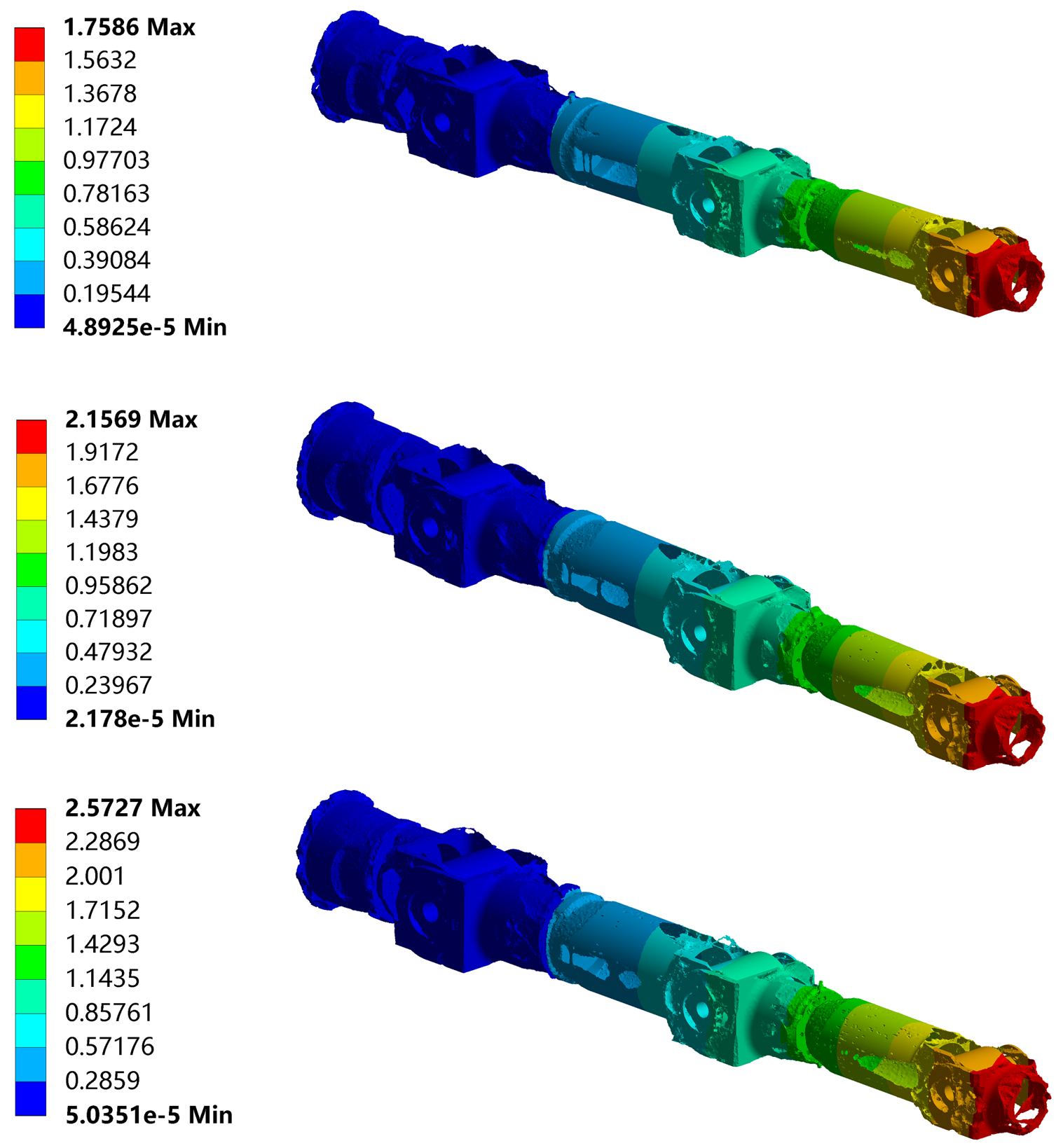}
    \caption{Deformation verification of optimized structures.
    Load condition 1 is shown as an example.}
    \label{fig9_arm_test}
\end{figure}

\begin{figure*}[t]
    \centering
    \includegraphics[width=6.8in]{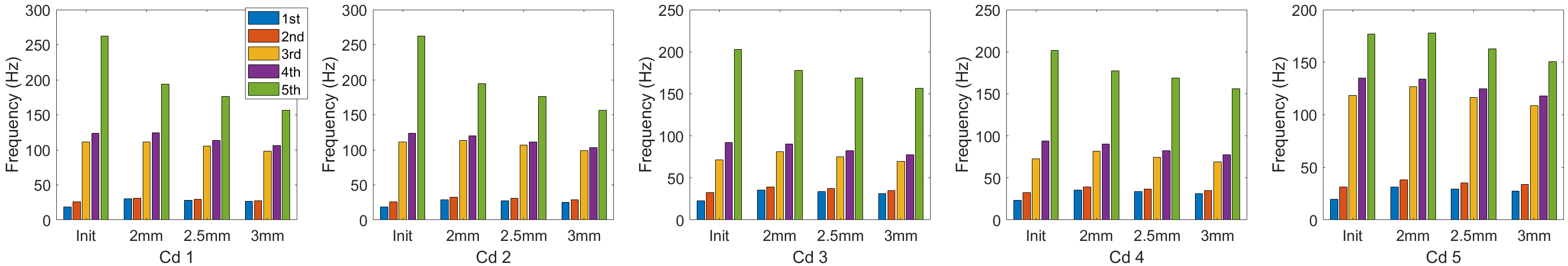}
\caption{Comparison of natural frequencies from the first mode to the fifth mode before and after optimization.}
    \label{fig10_arm_freq}
\end{figure*}

\section{Optimization of Robotic Arm}

Compared with the simple example above, optimizing a robot involves some additional challenges, 
such as multiple materials, multiple working conditions, and vibration performance. 
Our methods based on strain energy distribution are also convenient for addressing these issues. 
Specifically, expected strain energies under multiple working conditions can be calculated and used for multi-condition optimization. 
Part-level optimization based on the expected strain energy can also employ size optimization or material optimization 
instead of topological optimization. 
For example, by changing only the material of a part, its strain energy changes proportionally to the Young's modulus.
Moreover, the vibration performance is positively related to stiffness. 
If the vibration exceeds expectations, it is only necessary to reduce the expected strain energy and re-optimize. 
In addition, if the stiffness of the initial design is below the desired value, 
our method can also improve the stiffness while minimizing the weight. 

As an example to demonstrate how to deal with these issues, we consider a robotic arm with a typical design as shown in Figure~\ref{fig7_arm_sim0}. 
It has 6 DOFs (6 joint actuators connected by 7 links). 
The actuators are not optimized here. 
For simplicity, integrated joint modules from a general supplier~\cite{zeroerr} are used for the necessary evaluation of weight and dimensions. 
Specifically, to achieve a rated load of $5\,kg$, the shoulder, elbow, and wrist joints use modules 
with rated torques of $50\,N\,m$, $30\,N\,m$, and $10\,N\,m$, respectively. 
All 6 joint modules have a total weight of $8.08\,kg$. 
They are treated as ``rest'' in Equation~\ref{eq20}. 
The supporting structures include the 7 main links (Links 0--6, Figure~\ref{fig7_arm_sim0}) and some small structures (bearing pedestal, bearing cover). 
They are all initial designs without manual optimization. 
Assuming all structural parts are made of aluminum alloy, their total weight is $7.116\,kg$, which is $46.8\%$ of the total weight of the whole arm. 

Before optimization, a full-model FEA is built to analyze the stiffness. 
Usually, the worst static load conditions are considered in FEA. 
For a robotic arm, the maximum gravity loads are applied when the arm is raised to a horizontal pose. 
For each link, the maximum load in every direction should be considered. 
Through this analysis, we select as few as 5 poses (Figure~\ref{fig7_arm_sim0}) that contain the complete loads on each link (Table~\ref{tab1}).
In the full-model FEA, an end load of $5\,kg$ and gravity are applied. 
The results are shown in Figure~\ref{fig7_arm_sim0}.
The stiffness is relatively high except for Condition 1, because the initial design is solid without manual optimization to reduce weight.
Load condition 1 has a relatively large deformation approaching $3\,mm$, 
while the deformations of the other load conditions are all less than $1.5\,mm$.
This indicates that the initial design has weaknesses in some areas. 
We will therefore implement lightweight optimization considering multiple working conditions and improve the stiffness at the same time.

First, we specify the expected stiffness targets. 
Three cases are given for comparison: the expected maximum deformation is less than $2\,mm$, $2.5\,mm$, and $3\,mm$, 
which are markedly less than, slightly less than, and approximately equal to the maximum deformation of the initial design, respectively.
Then, from the full-model FEA of the initial design, the strain energies of the whole arm $U_{tot}$ and each part $U_i$ can be obtained. 
Since the 7 main link structures contribute most of the structural weight, 
we optimize only them and treat the other small structures and actuators as $U_{rest}$ in Equation~\ref{eq20}.
Based on the deformation of the initial design and the expected targets, 
the expected strain energy of the whole arm $U_{tot}^{ref}$ can be calculated by Equation~\ref{eq19}.
Based on Equation~\ref{eq22}, the expected strain energy of each part $U_i^{ref}$ can be obtained.
These calculations are performed for each load condition.
Taking the $2\,mm$ target as an example, the strain energy calculations are listed in Table~\ref{tab3}.

With the optimization objectives assigned above, we can find that some parts have a lower expected strain energy than the initial design,
which indicates they are weak points of the system and should be strengthened (Links 1 and 2, Table~\ref{tab3}).
In contrast, some parts may have a much higher expected strain energy than the initial design under some conditions, 
which indicates they are overly redundant in those cases and can be significantly lightened. 
We can then optimize each part to achieve $U_i\rightarrow U_i^{ref}$.

For each part, manual optimization, material optimization, topological optimization, and shape and size optimization can all be chosen. 
Here, we use material and topological optimization as examples. 
Topological optimization can hardly strengthen a part because it is primarily aimed at reducing material. 
However, if the constraint is too weak, topological optimization will generate a very thin structure.
We first apply material optimization to those parts for which $U_i^{ref} \le U_i$ or $U_i^{ref} \gg U_i$. 
Commonly used metallic materials are considered here (Table~\ref{tab2}).
They have similar specific stiffness, i.e., greater stiffness means greater weight, and thus none of them is clearly superior to the others.
If $U_i^{ref} \le U_i$, replace the material with one of higher stiffness and density. 
If $U_i^{ref} \gg U_i$, replace the material with one of lower stiffness and density. 
After material replacement, the part mass changes, and the expected strain energy $U_i^{ref}$ can be updated by Equation~\ref{eq22}. 
At the same time, the current strain energy $U_i$ of the part also changes and can be simply estimated as $(E_{old}/E_{new}) U_i$.
In this way, $U_i^{ref}$ and $U_i$ can be brought closer together.
By repeatedly changing materials, we can ensure that all parts satisfy $U_i^{ref} > U_i$.
If $U_i^{ref} < U_i$ persists, it indicates that the initial design of the part is too weak and needs to be further strengthened, 
or that additional materials need to be provided for selection. 
Here, feasible material selections are obtained through trial and error. 
The material optimization results are listed in Table~\ref{tab3}, and the topology optimization density distributions are shown in Figure~\ref{fig8_arm_TO}; the detailed strain energy calculations are also listed in Table~\ref{tab3} 
($2\,mm$ target as an example).

Then, we apply topological optimization to each part to achieve a lightweight design. 
$REB3$ elements are also used to equivalently simulate the boundary conditions of a part, connections to other parts, and the remote force. 
The 5 load conditions provide 5 deformation constraints. 
The minimum-weight design can be obtained when the 5 constraints are satisfied simultaneously. 
Obviously, not all constraints can reach their bounds, which means redundancy always exists in some load conditions. 
As discussed above, we use an additional iteration to obtain a better lightweight design (Figure~\ref{fig6_flow_chart}, Equation~\ref{eq25}). 
In total, only two rounds of part-level optimization for each part can provide an effective system-level optimization.
Figure~\ref{fig8_arm_TO} shows the topological optimization results of every optimized part. 
Less material is retained when $U_i^{ref}/U_i$ is larger.
The lightweight results are shown in Table~\ref{tab4}. 
For the $2\,mm$, $2.5\,mm$, and $3\,mm$ targets, 
the structural weight is reduced from $7.116\,kg$ to $4.266\,kg$, $3.298\,kg$, and $2.824\,kg$, respectively (based on the mesh-density results). 
The weight reduction is substantial.

To verify whether the system stiffness achieves the desired effect, we assemble the mesh models from topological optimization into the full-model FEA. 
As expected, the maximum deformations under every load condition are lower than the given targets ($2\,mm$, $2.5\,mm$, and $3\,mm$) (Table~\ref{tab4}). 
In particular, under load condition 1 (Figure~\ref{fig9_arm_test}), the maximum deformations decrease to the expected ranges. 
This indicates that lightweight optimization and stiffness improvement are achieved simultaneously. 
The stiffness of the initial design under the other load conditions was sufficient. 
After optimization, it decreases slightly but remains better than the given targets. 
This reduces the redundancy and results in a large reduction in weight.
The stiffness difference among the various load conditions becomes smaller, 
which means that localized weaknesses are mitigated and the isotropy of the system's mechanical properties is enhanced.
It should be noted that the maximum deformation is less than the given expected targets, 
which is also because $REB3$ elements make the connection interfaces weaker. 
Greater stiffness is obtained when the parts are assembled, which means the lightweight design is still relatively conservative, 
although the lightweight effect has already been good.

Vibration performance is also a crucial factor that must be considered in robotic design. 
Vibration sources from actuators or the external environment 
may induce structural vibration and lead to poor dynamic performance and reduced control accuracy. 
Vibration performance is related to the stiffness and mass of the whole model. 
It is difficult to map it directly to the strain energy.
FEA can analyze the modes, natural frequencies, and forced vibration amplitudes. 
In general, the greater the stiffness, the higher the natural frequency. 
The forced vibration is also influenced by the natural frequency and stiffness.
Therefore, if the vibration performance is unsatisfactory, it can be adjusted by changing the stiffness or mass.
The corresponding approach here is to change the given expected deformation target and perform a new optimization.
As shown in Figure~\ref{fig10_arm_freq}, the natural frequencies are compared from the first mode to the fifth mode. 
After optimization, the low-order frequencies are higher than those of the initial design, which is consistent with the stiffness improvement.
As the expected stiffness decreases, the natural frequencies also decrease.
In contrast, the high-order frequencies are lower than those of the initial design. 
This is probably due to the stiffness decrease under some load conditions. 
Overall, by changing the stiffness optimization objective, the vibration performance can be adjusted.

\section{Conclusions}

In this work, we proposed a new systematic method for the lightweight optimization of complex mechanical systems.
The method is based on a criterion that the specific strain energy should be uniformly distributed across the system 
when the system has the optimal stiffness-to-weight ratio.
By optimally assigning the strain energy of each part, the system-level optimization is decomposed into part-level optimization 
without loss of systematicness. 
In this way, any optimization method for individual parts can be selected for the part-level optimization, 
and thus this method is highly flexible and does not depend on specific optimization methods.
On one hand, it avoids system-level computational complexity; on the other hand, it retains systematic optimization.
Therefore, the method achieves effective weight reduction. 
Effective weight reduction can be achieved together with stiffness control, improvement, or maintenance. 
Compared with the existing systematic approach, the metamodel method~\cite{kim2016topology,wang2019optimal,hu2020multi},
our method is more flexible and requires lower computational cost.
As an example, we demonstrated how to implement the lightweight optimization of a robotic arm considering multiple load conditions and materials. 
The results showed that the structural weight after optimization decreases to a very low fraction, approximately $26\%$ of the total weight of the arm, 
while the system stiffness is still improved (Table~\ref{tab4}). 
The further potential for structural weight reduction is evidently small, and actuators then become the dominant contributor to the weight. 

Although the proposed method achieves substantial theoretical weight reduction, which is the main contribution of this work, 
there are still many practical issues to address. 
Redesigning and manufacturing a topology may introduce additional weight.
An ideal approach is to manufacture the parts directly by 3D printing, which can best preserve the lightweight effect.
However, there are usually various other practical requirements, such as manufacturing method, manufacturing costs, and appearance,
which impose additional manufacturing constraints.
How best to preserve the lightweight effect during manufacturing remains an open problem.
Topological optimization software usually provides manufacturing constraint options for these issues.
Adding manufacturing constraints to topological optimization will reduce the lightweight effect to some extent.
It should also be noted that the theoretical proof of the uniform specific strain energy criterion is strictly derived for serial-chain systems.
For more complex structural topologies, the criterion is supported by FEA validation rather than formal proof, and its generalization warrants further theoretical investigation.

On the other hand, there are also measures that can be taken to achieve further weight reduction. 
First, only 2 iterative steps are performed in the lightweight optimization (Table~\ref{tab4}).
More iterative steps can yield further weight reduction. 
However, the additional effect is limited. 
The second iterative step brings about only a $0.3\,kg$ decrease in total weight, 
and it can be predicted that a third step would bring only a $0.1\sim0.2\,kg$ decrease in total, which has a low cost-to-benefit ratio. 
Additional iterative steps are possible but seem unnecessary.
Second, we used $REB3$ elements to equivalently simulate the boundary conditions and connections for the part-level simplification.
$REB3$ elements are widely used, convenient, and available in various FEA software packages.
However, $REB3$ elements make the connection interface weaker than direct connections,
which causes the stiffness to be underestimated, and thus greater stiffness is obtained than expected from optimization.
Therefore, a larger expected deformation can be specified to obtain further weight reduction.
$REB2$ elements can also be used, but are not recommended because they make the interface stiffer.
Some research works have provided methods for defining the boundary conditions of parts within an assembly. 
Ref.~\cite{sha2020topology} used FEA of the assembly to obtain the displacement boundary conditions of each part 
and then performed topological optimization for each part. 
Ref.~\cite{liu2022topology} performed topological optimization for each individual part directly within the assembly model. 
These methods can serve as alternatives to $REB3$.

There is one more issue that warrants discussion.
Part-level optimization can flexibly use various methods.
Here, we use material optimization as an example because changes in material allow the strain energy changes to be easily estimated. 
Size optimization and manual optimization can also be chosen, as long as the resulting strain energy changes can be obtained. 
In practice, not all parts are suitable for topological optimization, such as small or geometrically simple parts, and other methods should then be considered. 
Composite materials are also an alternative. 
FEA support for composite materials is widely available in various software packages and can be used to optimize the strain energy of a part.
Similarly, some new manufacturing techniques, such as lattice materials, can also be incorporated through FEA within our systematic optimization framework.

Taken together, our method is a systematic approach, and therefore the optimization can be more effective.
Individual component optimization may address only specific issues or weaknesses in a system. 
In contrast, systematic optimization takes a holistic approach to identify and optimize all components of the system 
and can achieve a more comprehensive and efficient improvement in overall performance.

%%%%%%%%%%%%% Acknowledgements %%%%%%%%%%%%%
\footnotesize
% \section*{Acknowledgements}

%%%%%%%%%%%%%%   Bibliography   %%%%%%%%%%%%%%
\normalsize
\bibliography{main}

%%%%%%%%%%%%  Supplementary Figures  %%%%%%%%%%%%
% \clearpage

%%%%%%%%%%%% Supplementary Methods %%%%%%%%%%%%
\footnotesize
% \section*{Supplementary material}

%%%%%%%%%%%%%%%%   End   %%%%%%%%%%%%%%%%
%\end{multicols}  % Method B for two-column formatting (doesn't play well with line numbers), comment out if using method A
\end{document}